%% file: main.tex
\documentclass[preprint,5p,times,twocolumn]{elsarticle}

\usepackage{lineno,hyperref}
\modulolinenumbers[5]

\usepackage{verbatim}
\usepackage[linesnumbered,lined,boxed,commentsnumbered]{algorithm2e}
\usepackage{latexsym, graphicx, epsfig}
\usepackage{url}
\usepackage{color}
\usepackage{array}
\usepackage{float}
\usepackage{amsfonts,amsmath,amssymb}
\usepackage{verbatim}
\usepackage{subfigure}
\usepackage{setspace}
\usepackage{bbm}
\usepackage{bm}

\usepackage{graphicx}
\usepackage{caption}
\usepackage{mcode}
\DeclareMathOperator*{\argmin}{argmin} 

\newtheorem{remark}{Remark}[section]

\def\mb{\mathbf}

\begin{document}
\journal{Theoretical and Applied Mechanics Letters}

\title{Nonnegativity-Enforced Gaussian Process Regression}

\author[uiowa]{Andrew Pensoneault}
\ead{andrew-pensoneault@uiowa.edu}
\author[lehigh]{Xiu Yang\corref{correspondingauthor}}
\ead{xiy518@lehigh.edu}
\author[uiowa]{Xueyu Zhu\corref{correspondingauthor}}
\cortext[correspondingauthor]{Corresponding author}
\ead{xueyu-zhu@uiowa.edu}

\address[uiowa]{Department of Mathematics, University of Iowa, Iowa City, IA 52246, USA}
\address[lehigh]{Department of Industrial and Systems Engineering, Lehigh University, Bethlehem, PA 18015, USA}

\begin{abstract}
Gaussian Process (GP) regression is a flexible non-parametric approach 
to approximate complex models.
In many cases, these models correspond to processes with bounded physical properties.
Standard GP regression typically results in a proxy model which is unbounded for all temporal or spacial points, and thus leaves the possibility of taking on infeasible values. We propose an approach to enforce the physical constraints in a probabilistic way under the GP regression framework. In addition, this new approach reduces the variance in the resulting GP model.
\end{abstract}


\maketitle

\pagestyle{myheadings}
\thispagestyle{plain}


\input{ConstraintGP/Sections/Introduction.tex}
\input{ConstraintGP/Sections/GP.tex}
\input{ConstraintGP/Sections/CGP.tex}
\input{ConstraintGP/Sections/Examples.tex}
\input{ConstraintGP/Sections/Summary.tex}
\bibliographystyle{abbrv}
\bibliography{Ref.bib}


\end{document}

%% file: ConstraintGP/Sections/Introduction.tex
\section{Introduction}
\label{sec:Intro}

In many applications, evaluating a computational model can require significant computational resources and time. One approach to address this problem is to build a surrogate model with statistical emulators such as Gaussian Processes (GP) regression \cite{sacks1989}.
We aim to design surrogate models that have low approximation error, and satisfy meaningful bounds on some physical properties. However, no such information is encoded in the standard GP regression method. Therefore, it can produce infeasible predictions.
 
Incorporating physical information in GP has been explored in many works of literature. 
For example, it is demonstrated in \cite{Salzmann2010} that the mean prediction of a GP model satisfies a set of linear equality constraints provided the training data satisfy these constraints. A similar result holds for quadratic equality constraints under a transformation of the parameterization. Alternatively, linear equality constraints can be enforced by modeling the process as a transformation of an underlying function and imposing the constraints on that transformation \cite{Jidling2017}. Moreover, physical information in the form of differential operators can be incorporated in GP models \cite{schober2014probabilistic,raissi2018numerical, Yang2018, yang2018physics}.

Incorporating inequality constraints in a GP is more difficult, as the underlying process conditional on the constraints is no longer a GP \cite{Maatouk2017}. To address this problem, several different approaches have been explored. The approach in \cite{Abrahamsen2001} enforces inequality constraints at several locations and draws approximate samples from the predictive distribution with a data augmentation approach.  Linear inequality functional (such as monotonicity) are enforced via virtual observations at several location within \cite{wang2016estimating,Agrell2019,golchi2013monotone,riihimaki2010gaussian,DaVeiga2012}. In \cite{Agrell2019}, it is shown that when linear inequality constraints are applied to a finite set of points in the domain, the process conditional on the constraints is a compound GP with a truncated Gaussian mean. In \cite{Maatouk2017,lopez2019approximating, MAATOUK201538},  linear inequality constraints are enforced on the entire domain instead of a finite set of points by making a finite-dimensional approximation of the GP and enforcing the constraints through the choice of the associated approximation coefficients. 

In this work, we focus on enforcing non-negativity in the GP model. This is a requirement for many physical properties, e.g., elastic modulus, viscosity, density, and temperature.
We propose to impose this inequality constraint with high probability via selecting a set of constraint points in the domain and imposing the non-negativity on the posterior GP at these points. In addition to enforcing non-negativity, this approach improves accuracy and reduces uncertainty in the resulting GP model. 

The paper is organized as follows. We review the standard GP regression framework in Section~\ref{sec:GP}, present our novel approach to enforce non-negativity in GP regression in Section~\ref{sec:CGP}, and provide numerical examples in Section~\ref{sec:NumExample}.

%% file: ConstraintGP/Sections/GP.tex
\section{Gaussian Process Regression}
\label{sec:GP}
We introduce the framework for GP regression based on the descriptions in~\cite{Rasmussen2004}. Assume we have $\bm y= (y^{(1)},y^{(2)},...,y^{(N)})^T$ as the values of the target function, where $y^{(i)}\in\mathbb{R}$ are observations at locations $\mathbf{X}=\{x^{(i)}\}^N_{i=1}$ where $x^{(i)}$ are $d$-dimensional vectors in the domain $D\subseteq\mathbb{R}^d$. 
We aim to use a GP $Y(\cdot,\cdot):D\times\Omega\to \mathbb{R}$ to approximate the underlying target function. 
Typically, $Y(x)$ is denoted as
\begin{align}
Y(x)\sim\mathcal{GP}(\mu(x),\mathcal{K}(x,x')),
\end{align}
where $\mu(\cdot):D\to\mathbb{R}$ and $\mathcal{K}(\cdot,\cdot):D\times D\to\mathbb{R}$ are the associated mean function and covariance function, i.e 
\begin{align}
    \mu(x) &= \mathbb{E}(Y(x)),\\
    \mathcal{K}(x,x')  &= \mathbb{E}(Y(x)-\mu(x))(Y(x')-\mu(x')).
\end{align}
A widely used kernel is the standard squared exponential covariance kernel with an additive independent identically distributed Gaussian noise term $\epsilon$ with
variance $\sigma^2_n$:
\begin{align}
\label{eq:kernel}
\mathcal{K}(x,x') = \sigma^2 \exp\left(\frac{-|x-x'|_2^2}{2l^2}\right)+\sigma_n^2\delta_{x,x'}.
\end{align}
where $\delta_{x,x'}$ is a Kronecker delta fuction,
$l$ is the length-scale, and $\sigma^2$ is the signal variance.
In general, by assuming zero mean function $\mu(x)\equiv 0$, we use $\theta=(\sigma,l, \sigma_n)$ to denote the hyperparameters, and they are determined based on the training data.

Define $K = [\mathcal{K}(x^{(i)},x^{(j)})]_{ij}$, $\boldsymbol{\mu}= (\mu(x^{(1)}),...,\mu(x^{(N)}))^T$, and $k(x')=[\mathcal{K}(x^{(i)},x')]_i-\sigma_n^2\delta_{x^{(i)}, x'}$. The posterior predictive distribution of the output  $y^*$ given the training set is $Y(x^*)|x^*,\mb{y},\textbf{X}\sim\mathcal{N}(y^*(x^*),s^2(x^*))$, where 
\begin{align}
    y^*(x^*) &= \mu(x^*) + k(x^*)^TK^{-1}(\boldsymbol{y}-\bm\mu),\\
    s^2(x^*) &= \sigma^2(x^*) -k(x^*)^TK^{-1}k(x^*). 
\end{align}
One approach to determine the hyperparameters $\theta$ is to minimize the negative marginal log-likelihood \cite{Rasmussen2004}:
\begin{align}\label{eq:loglike}
-\log(p(\boldsymbol{y}|\textbf{X},\theta)) = \frac{1}{2}\left[(\boldsymbol{y}-\boldsymbol{\mu})^TK^{-1}(\boldsymbol{y}-\boldsymbol{\mu})+\log|K|+N\log(2\pi)\right].
\end{align}

%% file: ConstraintGP/Sections/CGP.tex
\section{Gaussian Process with Constraint}
\label{sec:CGP}
In particular, we enforce the non-negativity in the quantity of interest. We minimize the negative marginal log-likelihood function in Eq.~\eqref{eq:loglike} while requiring that the probability of violating the constraints is small. More specifically, for $0<\eta\ll 1$, we impose the following constraint:
\begin{align}
P\big((Y(x)|x,y,\mathbf{X})<0\big)\leq\eta \text{ for all }x\in D.
\end{align}
This differs from other methods in the literature, which enforce the constraint via truncated Gaussian assumption \cite{Maatouk2017}, or  use a bounded likelihood function and perform inference based on the Laplace approximation and expectation propagation \cite{jensen2013bounded}. In contrast, our method retains the Gaussian posterior of standard GP regression, and only requires a slight modification of the existing cost function. As $Y(x)|x,y,\mathbf{X}$ follows a Gaussian distribution, this constraint can be rewritten in terms of the posterior mean $y^*$ and posterior standard deviation $s$:
\begin{align}
y^*(x)+\Phi^{-1}(\eta)s(x)\geq 0  \text{ for all }x\in D,
\end{align}
where $\Phi^{-1}$ is the inverse cumulative density function (CDF) of a standard Gaussian random variable. 
In this work, we
set $\eta=2.2\%$ for demonstration purpose, and consequently $\Phi^{-1}(\eta) = -2$, i.e., two standard deviations below the mean is still non-negative. Therefore, we minimize the negative log-likelihood cost function subject to constraints on the posterior mean and standard deviation:
\begin{align}
&\argmin_{\theta} \quad -\log(p(\boldsymbol{y}|\textbf{X},\theta)) \\
& \textrm{s.t.}  \quad 
y^*(x)-2s(x)\geq 0 \text{ for all }x\in D.
\label{eq:p_con}
\end{align}
We note that \eqref{eq:p_con} is a functional constraint and thus can be difficult to enforce. Instead, we enforce \eqref{eq:p_con} on a set of constraint points $\mb{X}_c=\{x_c^{(i)}\}_{i=1}^m$. Of note, these constraint points play similar roles as the aforementioned virtual observations \cite{wang2016estimating,Agrell2019,golchi2013monotone,riihimaki2010gaussian,DaVeiga2012}.  

Meanwhile, in practice, a heuristic on the distance of the posterior mean of the GP from the training data is applied to stabilize the optimization algorithm, as such to guarantee that 
it results in a model that fits measurement data. Subsequently, to obtain the constrained GP, we solve the following constrained minimization problem:
\begin{align}
&\argmin_{\theta} \quad -\log(p(\boldsymbol{y}|\textbf{X},\theta)) \\
\textrm{s.t.} \quad & 0\leq y^*(x_c^{(i)})-2s(x_c^{(i)}),\text{ for all } i=1,...,m.\\
 \quad & 0\leq  \epsilon-|y^{(j)}-y^*(x^{(j)})|,\text{ for all }j=1,...,n. \label{consistency-const}
\end{align}
where $\epsilon>0$ is chosen to be sufficiently small. In the this paper, we set $\epsilon = 0.03$. The last constraint is chosen so that the given solution fits the data sufficiently well.

We remarked that compared with unconstrained optimization,  constrained optimization is in general more computationally expensive \cite{box1965new}. However, if non-negativity approximation of the target function is crucial for the underlying applications, one may weigh less on the efficiency in order to get more reliable and feasible approximation within the computational budget.   


%% file: ConstraintGP/Sections/Examples.tex
\section{Numerical Examples}
\label{sec:NumExample}
In this section, we present numerical examples to illustrate
the effectiveness of our method. We measure the relative $l_2$ error between the posterior mean $y^*$ and the true value of the target function $f(x)$ over a set of test points $\mb{X}_T =\{x_T^{(i)}\}_{i=1}^{N_T}$: 
\begin{align}
E = \sqrt{\frac{\sum_{i=1}^{N_T}\big(y^*(x_T^{(i)})-f(x_T^{(i)})\big)^2}{\sum_{i=1}^T f(x_T^{(i)})^2}}.
\end{align}
For the examples below, we use $N_T = 1000$ equidistant test points over the domain $D$. We use the standard squared exponential covariance kernel as well as a zero prior mean function $\mu(x)=0$.
We solve the unconstrained log-likelihood minimization problem in MATLAB using the GPML package \cite{Rasmussen2010}. For the constrained optimization, we use the \mcode{fmincon} from the MATLAB Optimization Toolbox based on the built-in interior-point algorithm \cite{MATLAB:2019a}. 
\begin{remark}
If the method results in convergence to an infeasible solution, the optimization  is performed again with another random initial guess (with astandard Gaussian noise added to the base initial condition - $\theta_0 = (\log(l),\log(\sigma),\log(\sigma_n))=(-3,-3,-10)$.
\end{remark}

\subsection{Example 1} Consider the following function:
\begin{align*}
f(x)=\frac{1}{1+(10x)^4}+\frac{1}{2}e^{-100(x-\frac{1}{2})^2},\text{ }x\in[0,1].
\end{align*}
For our tests on this example, 
the training point set 
is $$\{x^{(i)}\}_{i=1}^{7}= \Big\{\frac{j-1}{5}+\epsilon_j\Big\}_{j=1}^{6}\cup\Big\{\frac{1}{2}\Big\},$$ where $\epsilon_j\sim\mathcal{N}(0,0.03^2)$ for $j=2,3,4,5$ and $\epsilon_1=\epsilon_6=0$. We chose $m=30$ equidistant points over the domain as our constraint points. 

\begin{figure}[h!]%

\centering
\subfigure[Unconstrained GP]{%
\includegraphics[height=.33\linewidth]{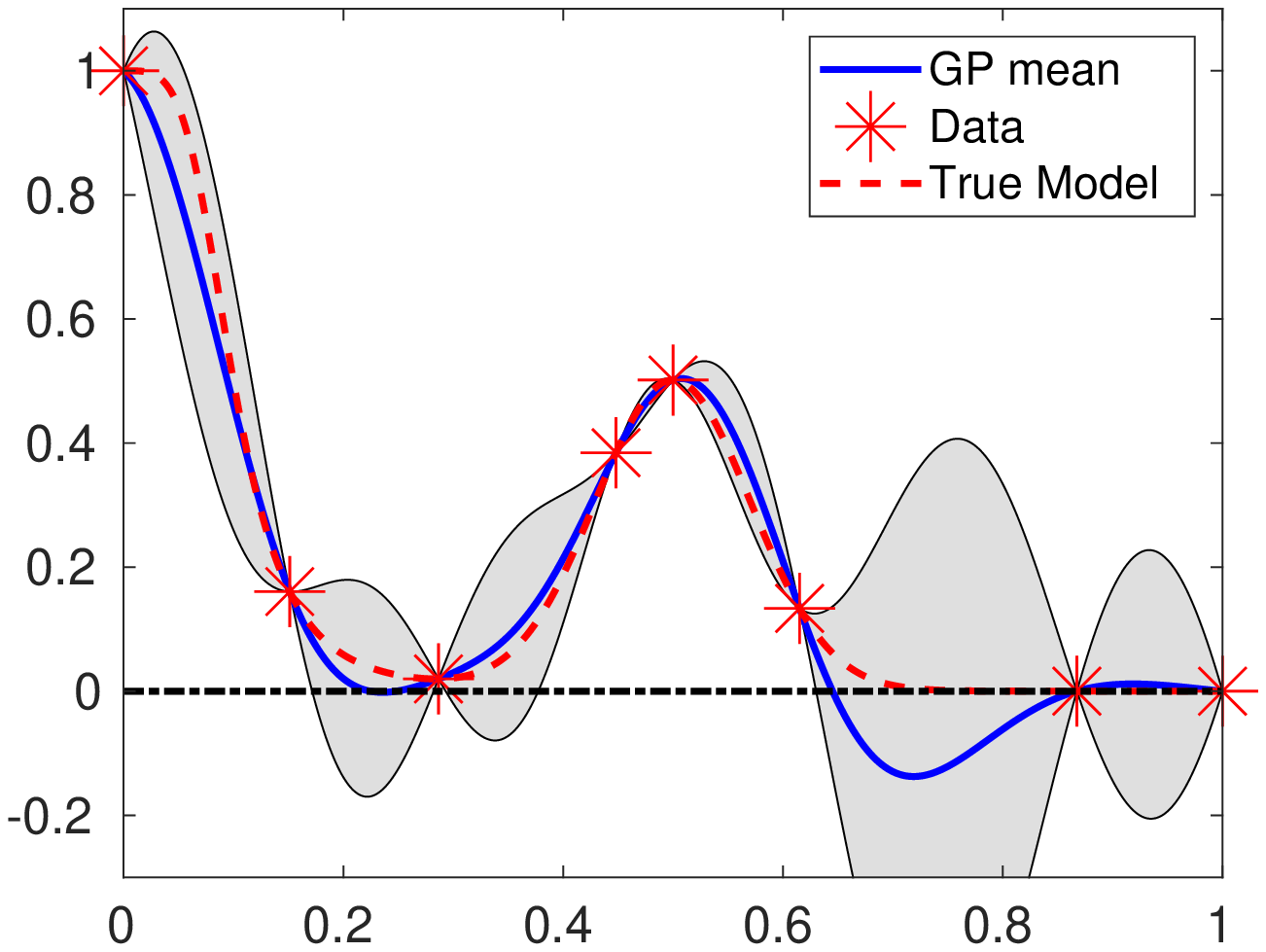}}
\quad
\subfigure[Constrained GP]{%
\includegraphics[height=.33\linewidth]{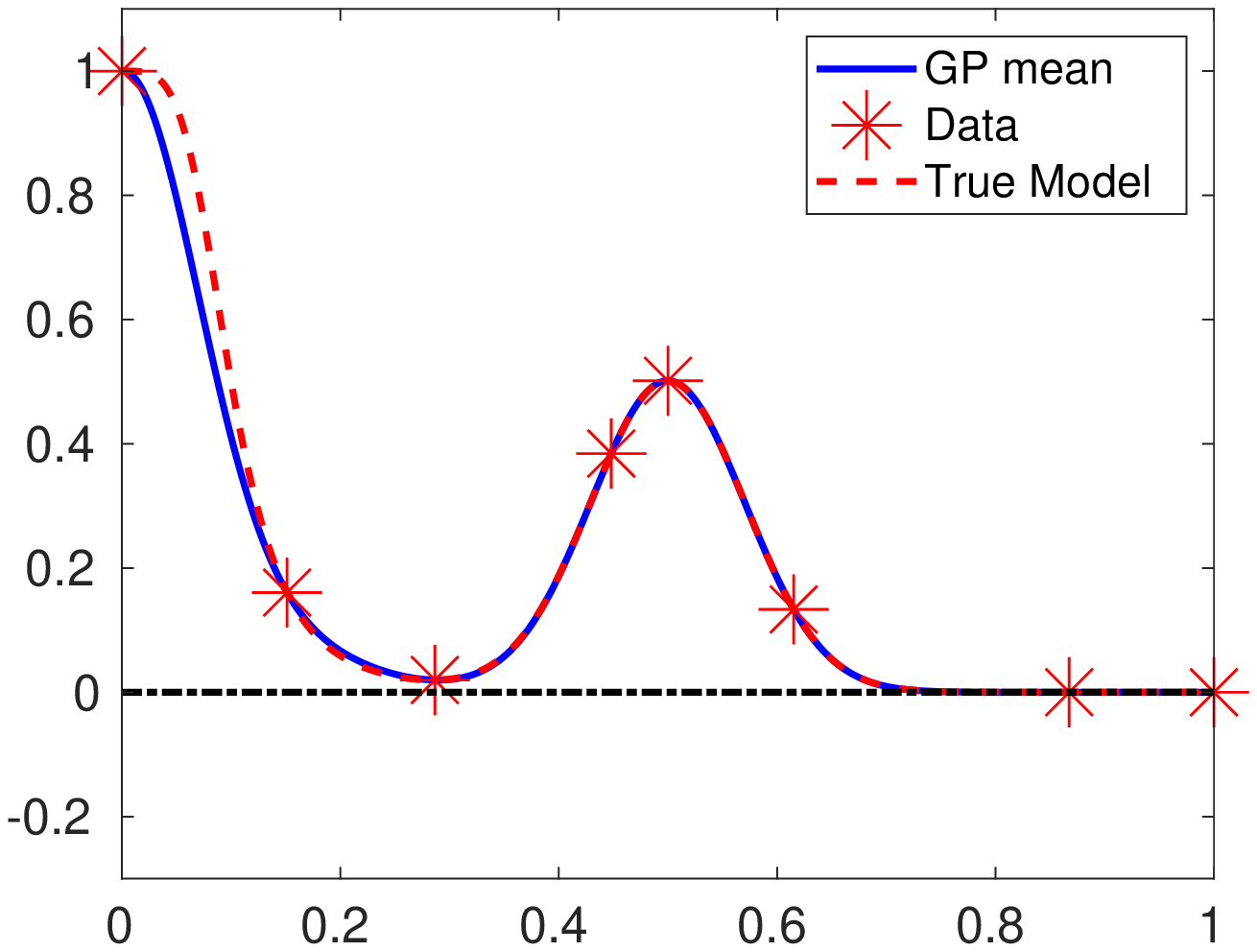}}%
\caption{The posterior mean and the corresponding $95\%$ confidence interval of the GP models in example 1. (a) Unconstrained GP. (b) Constrained GP.}
\label{fig:samp1}
\end{figure}

Figure \ref{fig:samp1} (a) shows the posterior mean of the unconstrained GP with $95\%$ confidence interval. It can be seen that on $[0.65,0.85]$, the posterior mean violates the non-negativity bounds with a large variance. In contrast, the posterior mean of the constrained GP in these regions no longer violates the constraints, as shown in Figure \ref{fig:samp1} (b). Besides, the confidence interval is reduced dramatically after the non-negativity constraint is imposed. 

\begin{figure}[h!]%
\centering
\subfigure[Relative Error]{%
\includegraphics[height=.33\linewidth]{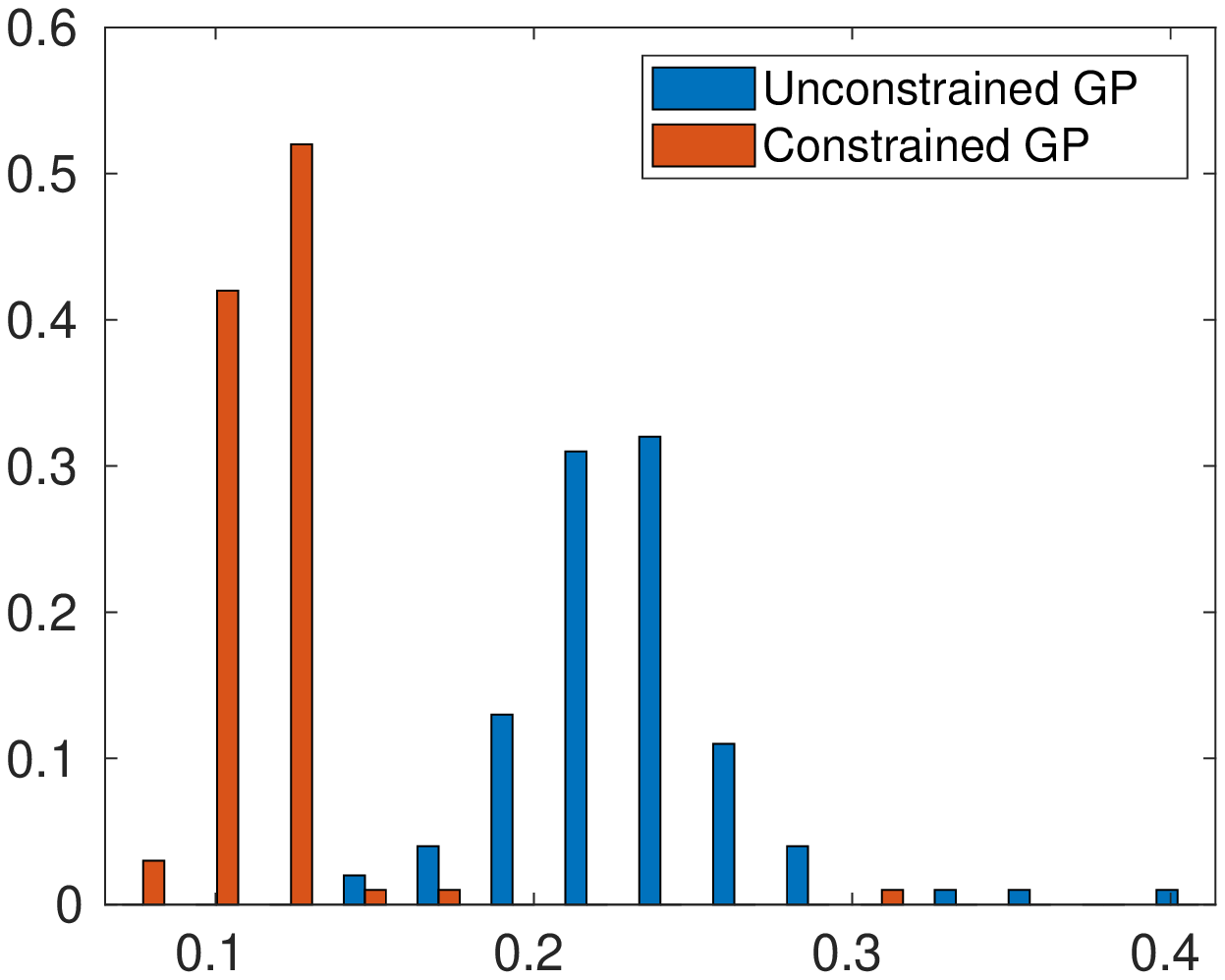}}
\quad
\subfigure[Percentage of Violation]{%
\includegraphics[height=.33\linewidth]{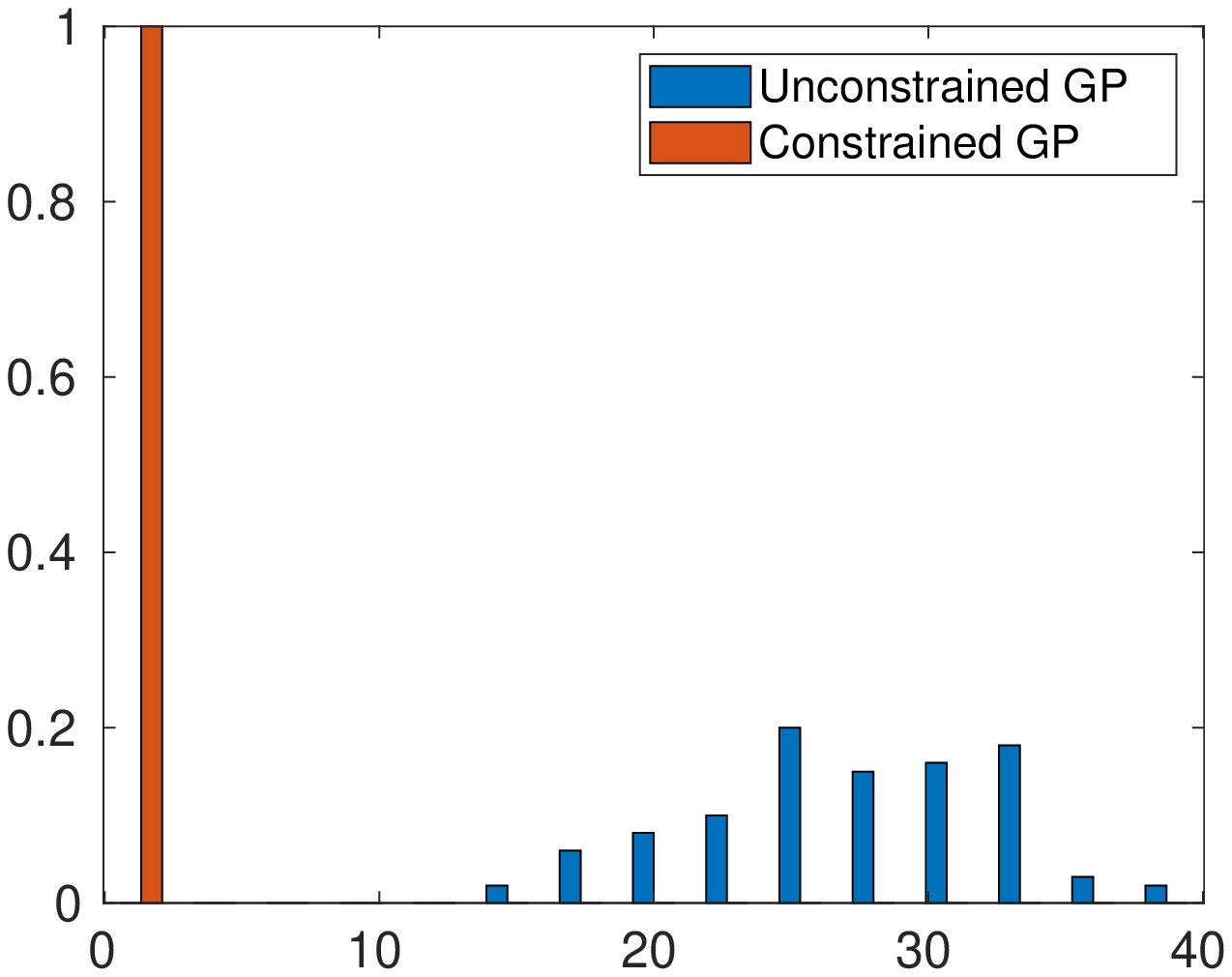}}
\caption{(a) The normalized histogram associated with the $l_2$ relative error between the GP mean and the true function over the test set based on 100 different training sets. (b) The normalized histogram associated with the percentage of the posterior mean over test points that violate the non-negativity constraint.}
\label{fig:hist1}
\end{figure}

To illustrate the robustness of the algorithm, we repeat the same experiment on 100 different training data sets as in \cite{DaVeiga2012}. 
Figure \ref{fig:hist1} (a) illustrates the distribution of the relative $l_2$ error over the 100 trials. It is clear that incorporating the constraint tends to result in a lower relative error in the posterior mean statistically. Figure \ref{fig:hist1} (b) compares the percentage of the posterior mean over the test points that violate non-negativity constraint over the 100 trails. There is a large portion of the posterior mean by the unconstrained GP that violates the non-negativity, while the constrained GP preserves the non-negativity very well.   

\subsection{Example 2}
Consider the following function:
\begin{align*}
f(x)&=\frac{1}{100} + \frac{5}{8}(2x-1)^4\big((2x-1)^2+4\sin(5\pi x)^2\big),\text{ }x\in[0,1].
\end{align*}
 We train our constrained and unconstrained GP models 
 over 14 training points at locations: 
 \begin{equation*}
     \{x^{(i)}\}_{i=1}^{14}= \Big\{\frac{j-1}{11}+\epsilon_j\Big\}_{j=1}^{12}\cup\{.075,.925\},
 \end{equation*}
 where $\epsilon_j\sim\mathcal{N}(0,0.03^2)$ for $j=2,...,11$ and $\epsilon_1=\epsilon_{12}=0$. We choose $m=31$ equidistant  points  in the domain as our constraint points.
 
\begin{figure}[h]%
\centering
\subfigure[Unconstrained GP]{%
\includegraphics[height=.33\linewidth]{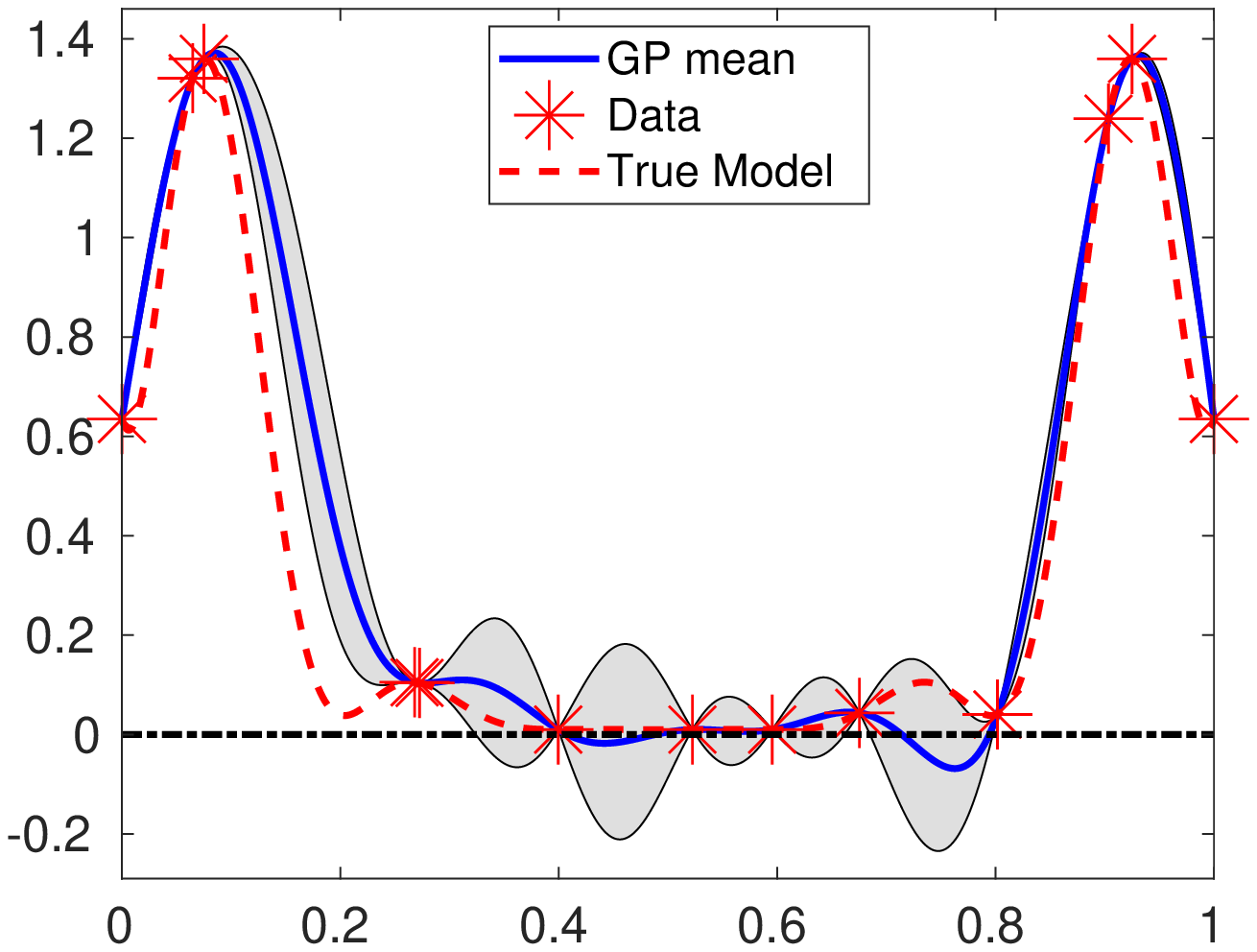}}
\qquad
\subfigure[Constrained GP]{%
\includegraphics[height=.33\linewidth]{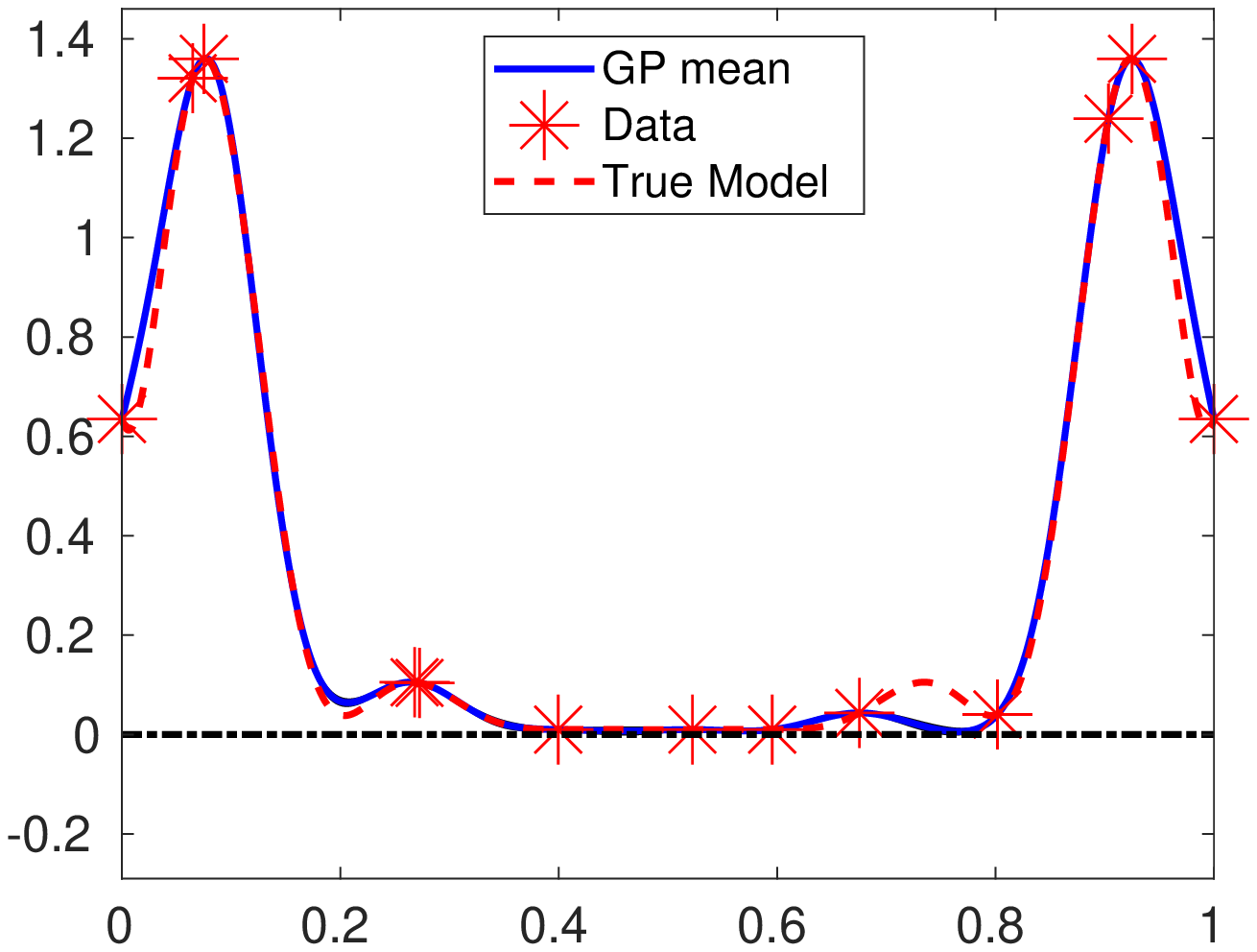}}%
\caption{The posterior mean and the corresponding $95\%$ confidence interval of the GP models in example 2. (a) Unconstrained GP. (b) Constrained GP.}
\label{fig:samp2}
\end{figure}

Figure \ref{fig:samp2} (a) shows a $95\%$ confidence interval around the posterior mean of the unconstrained GP. Notice that the posterior mean is less than zero near neighborhoods of $0.8$.
In contrast, the constrained GP doesn't violate the constraints as shown in Figure \ref{fig:samp2} (b). The confidence interval of the posterior mean is also much narrower, which illustrates the advantage of incorporating the constraints.

\begin{figure}[h]%
\centering
\subfigure[Relative Error]{%
\includegraphics[height=.33\linewidth]{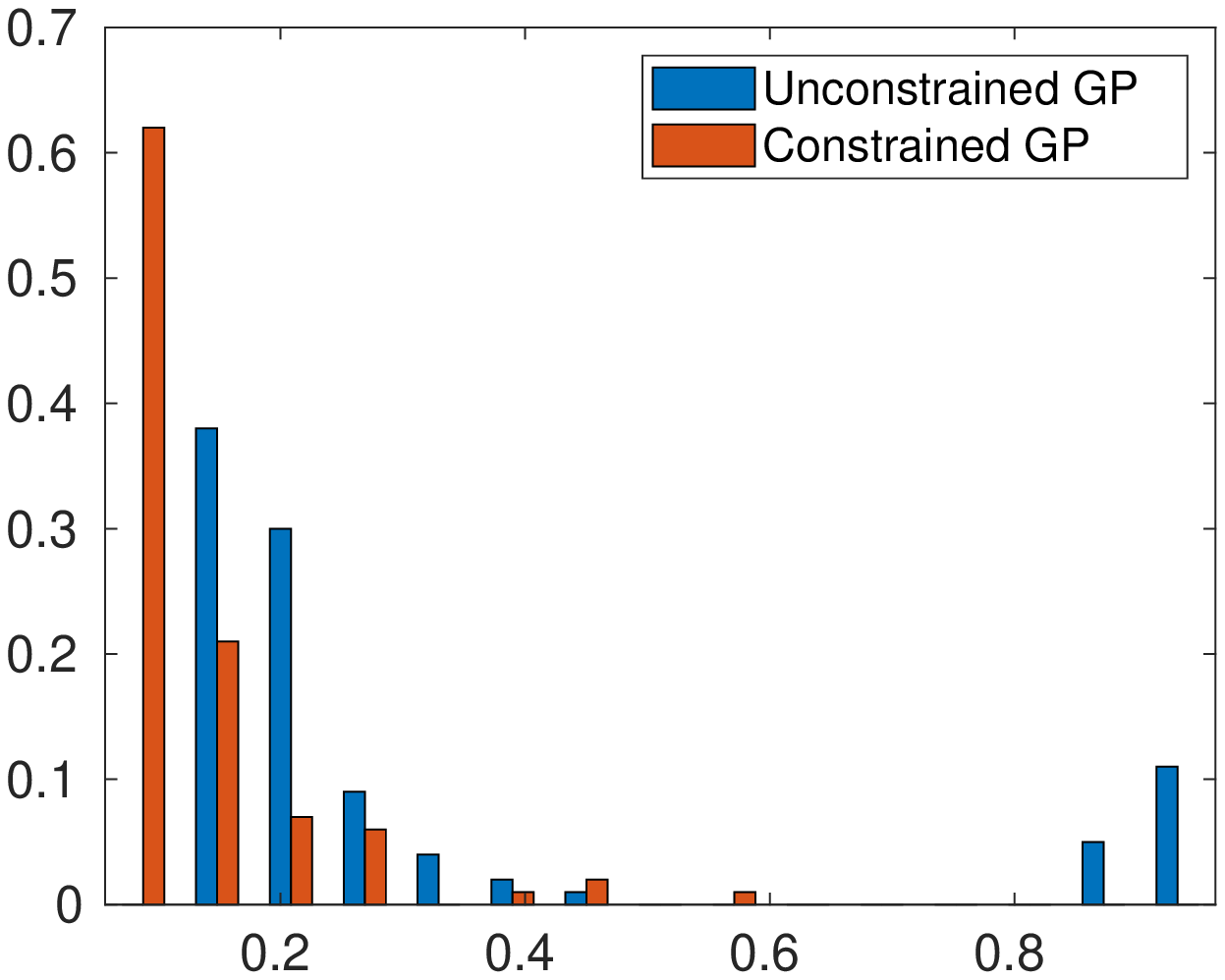}}
\quad
\subfigure[Percentage of Violation]{%
\includegraphics[height=.33\linewidth]{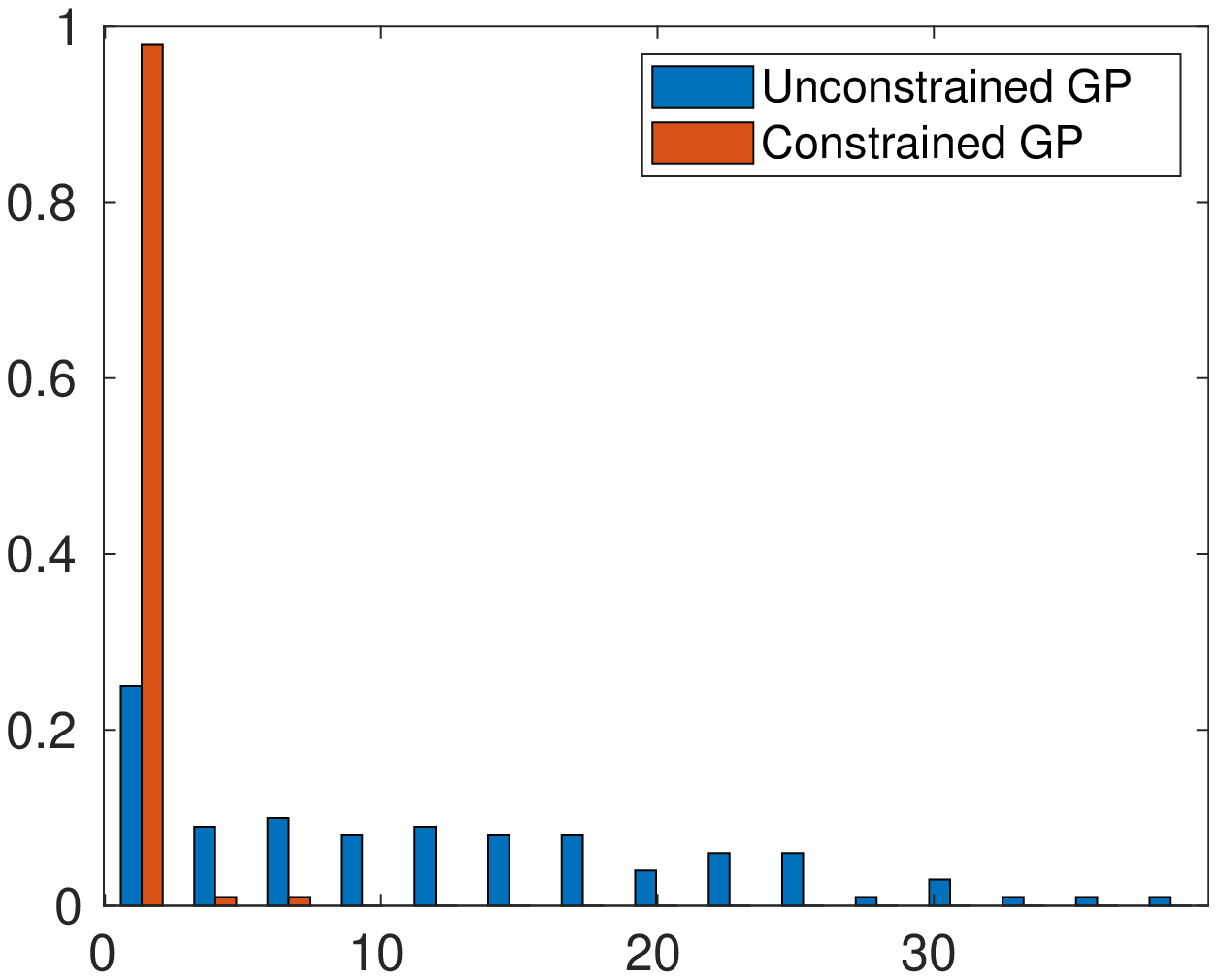}}%
\caption{(a)The normalized histogram associated with the $l_2$ relative error between the GP mean and the true function over the test set based on 100 different training sets. (b) The normalized histogram associated with the percentage of the posterior mean over test points which violate the non-negativity constraint.} 
\label{fig:hist2}
\end{figure}

Again, to show the robustness of the algorithm, we repeat the same experiment on 100 trials. Figure \ref{fig:hist2} (a) shows the relative $l_2$ error over 100 trials. The constrained GP has a histogram more heavily weighted towards lower relative error in the posterior mean, compared to the unconstrained GP. Figure \ref{fig:hist2} (b) shows that the posterior mean of the unconstrained GP  violates the non-negativity condition more frequently. 



\subsection{Example 3} 
The Korteweg-de Vries (KdV) \cite{Schalch2018} equation can be used to describe the evolution of solitons, which are characterized by the following properties: 1) invariant shape; 2)  
approaches a constant as $t\to \infty$; 3) strong interactions with other solitons.
We consider the KdV equation in the following form
\[u_t(x,t)-6u(x,t)u_x(x,t)+u_{xxx}(x,t)=0.\]
Under several assumptions on the form of $u$, an analytic solution can be found. For the case of two solitons, 
a (normalised) solution can be found in \cite{Schalch2018}:
\begin{equation}
    u(x,t) = \frac{12(3+4 \cosh(2x-8t)+\cosh(4x-64t))}{8[3 \cosh(x-28t)+\cosh(3x-36t))]^2}.
\end{equation}
For this equation, $u(x,t)>0$ for all $x,t\in\mathbb{R}$, we aim to approximate $u(x,-1)$ using GP. \\

We train our constrained and unconstrained GP model based on $13$ training points at locations: 
 \begin{equation*}
     \{x^{(i)}\}_{i=1}^{13}= \Big\{-10+15\frac{j-1}{10}+\epsilon_j\Big\}_{j=1}^{11}\cup\{-1.4,-8.4\},
 \end{equation*}
 where $\epsilon_j\sim\mathcal{N}(0,0.3^2)$ for $j=2,...,10$ and $\epsilon_1=\epsilon_{11}=0$. We choose $m=40$ equidistant points in the domain as our constraint points.\\
 
\begin{figure}[]%
\centering
\subfigure[Unconstrained GP]{%
\includegraphics[height=.33\linewidth]{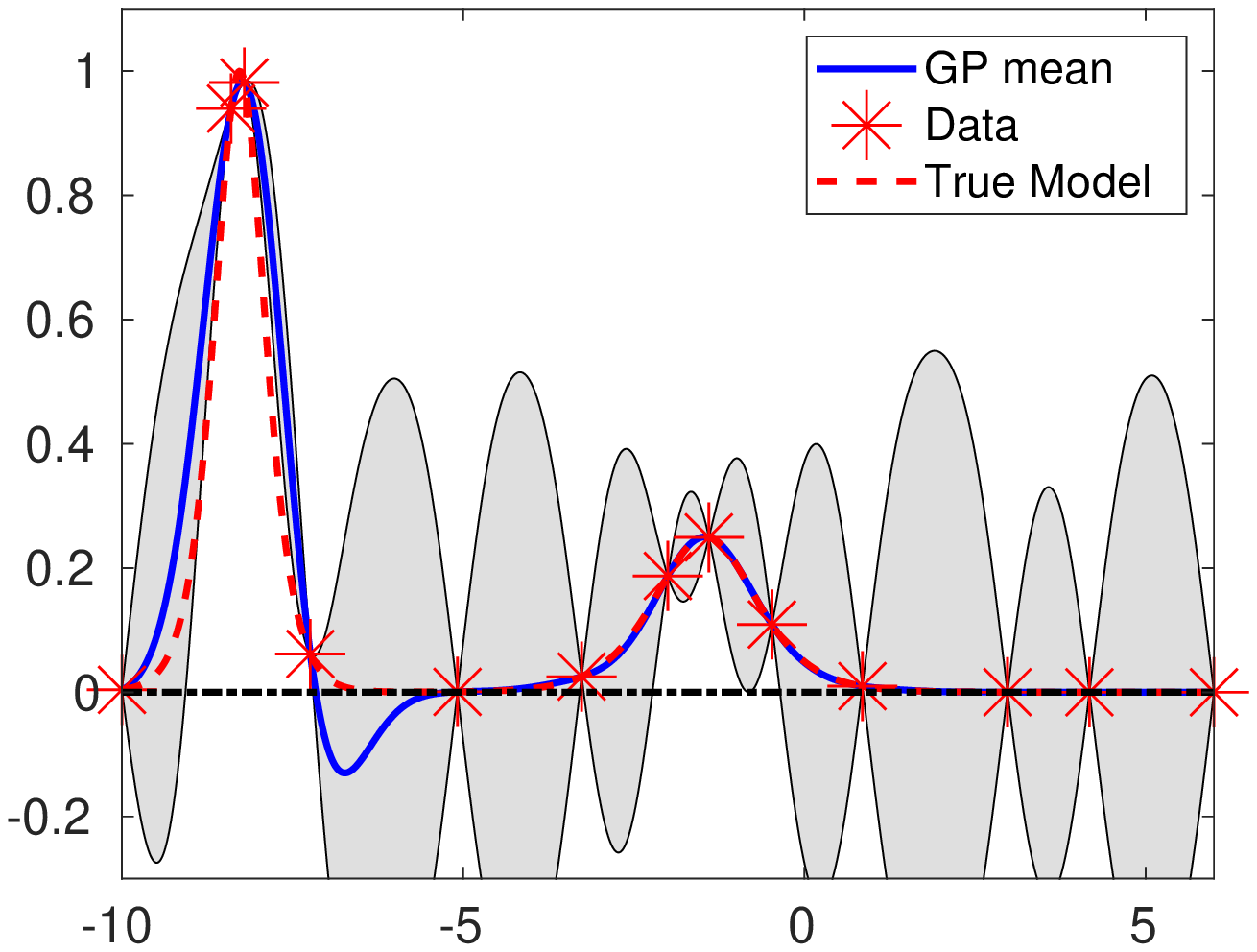}}
\quad
\subfigure[Constrained GP]{%
\includegraphics[height=.33\linewidth]{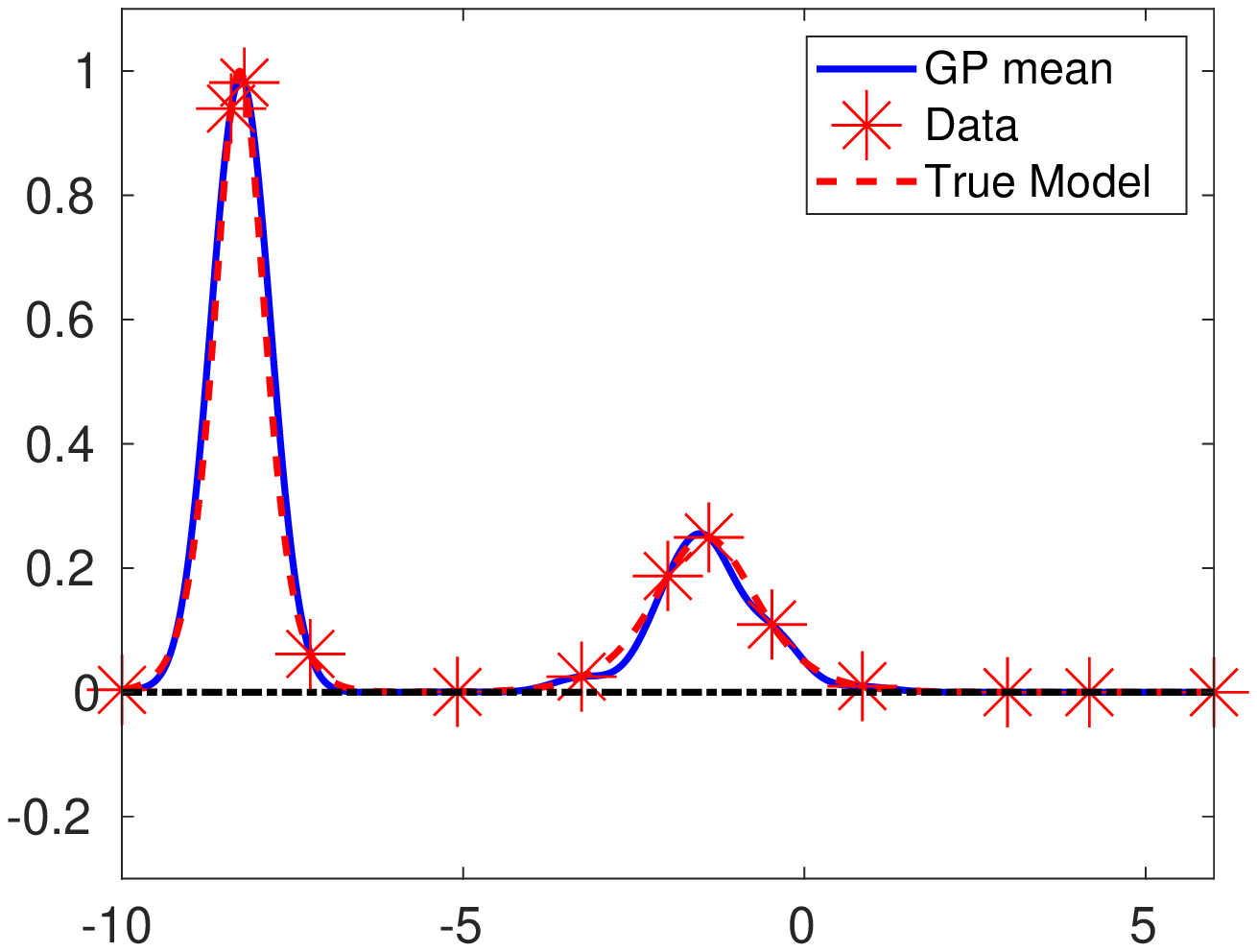}}
\caption{The posterior mean and the corresponding $95\%$ confidence interval of GP models approximating the two-soliton interacting system at $t=-1$ for a set training data set. (a) Unconstrained GP. (b) Constrained GP.}
\label{fig:samp3}
\end{figure}

\begin{figure}[]%
\centering
\subfigure[Relative Error]{%
\includegraphics[height=.33\linewidth]{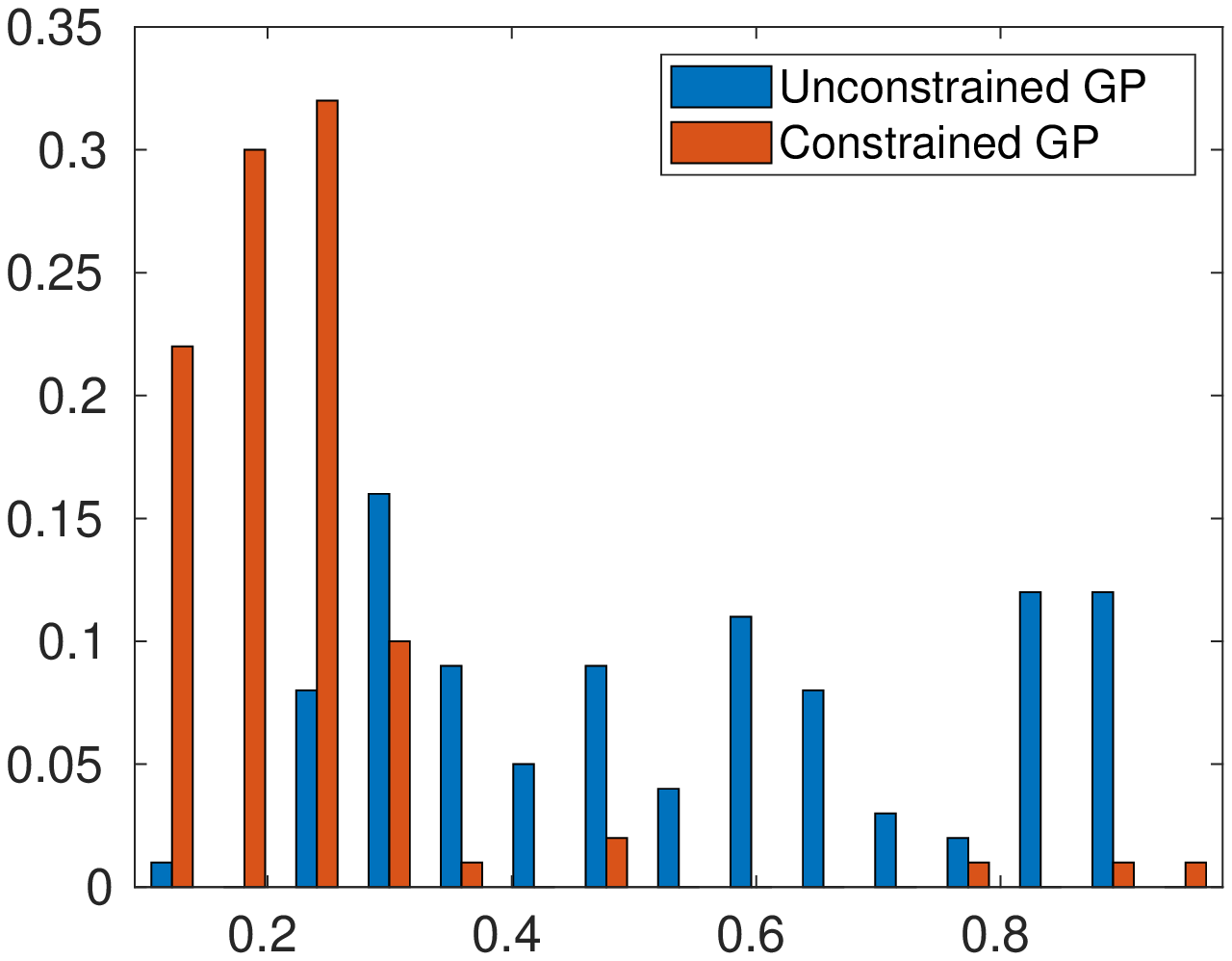}}
\quad
\subfigure[Percentage of Violation]{%
\includegraphics[height=.33\linewidth]{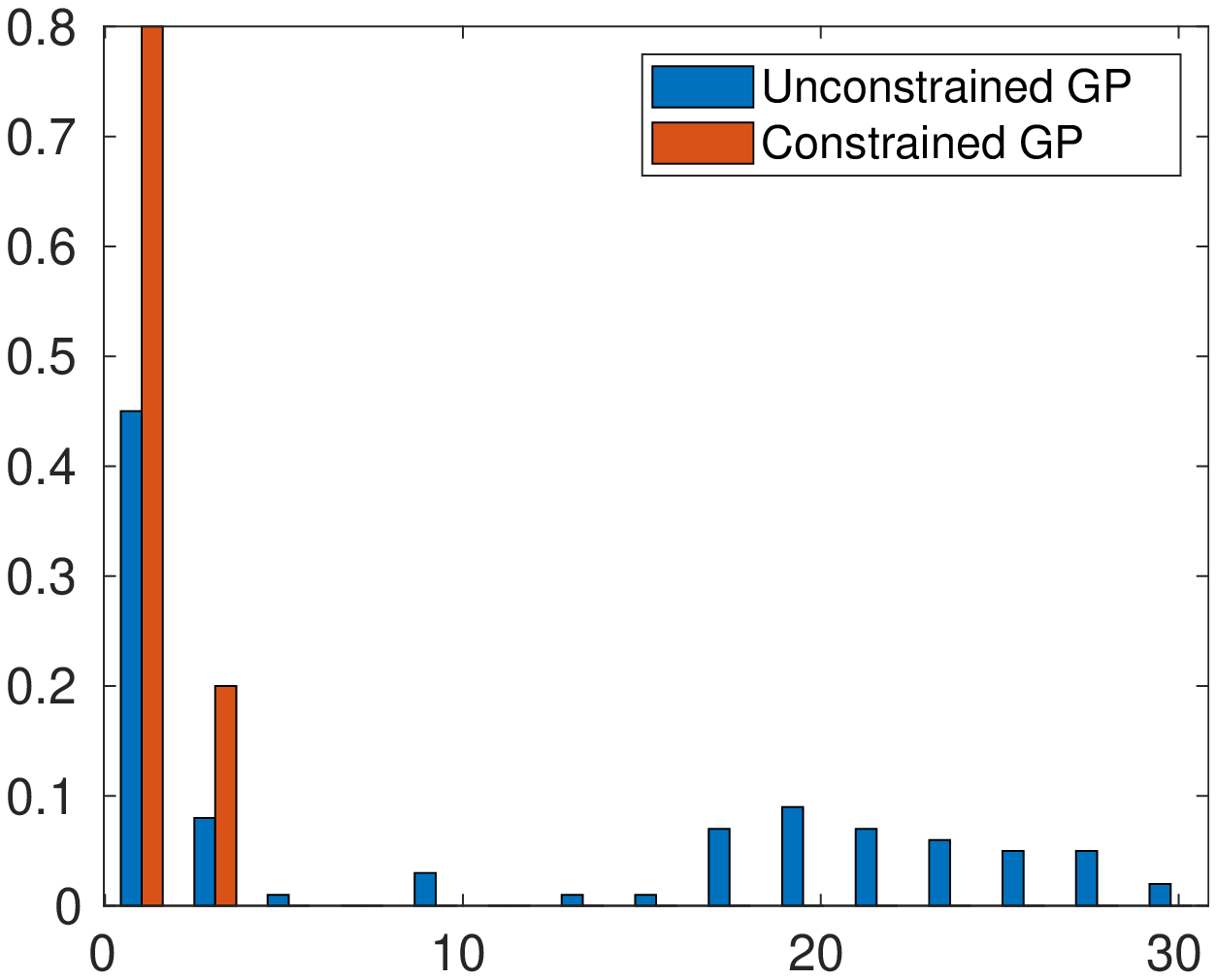}}%
\caption{(a) The normalized histogram associated with the $l_2$ relative error between the GP mean and the true function over the test set based on 100 different training sets. (b) The normalized histogram associated with the percentage of the posterior mean over test points that violate the non-negativity constraint.} 
\label{fig:hist3}
\end{figure}

 As can be seen in Fig \ref{fig:samp3}, the unconstrained GP violates non-negativity around $x=-7$, which is avoided in the constrained GP. More importantly, the confidence interval of the resulting GP is dramatically reduced by imposing non-negativity constraint. In addition, Fig. \ref{fig:hist3} (a) shows that the relative error is significantly reduced when we incorporate the non-negativity information. Of note, in this case, because the majority of the test points are near zero, the relative error is much more sensitive to approximation errors in these regions. Fig. \ref{fig:hist3} (b) illustrates that the constrained GP preserves the non-negativity with very high probability while the unconstrained GP violates the non-negativity much more frequently.

%% file: ConstraintGP/Sections/Summary.tex
\section{Summary}

In this paper, we propose a novel method to enforce the non-negativity constraints on the GP in the probabilistic sense. 
This approach not only reduces the difference between the posterior mean and the ground truth, but significantly lowers the variance, i.e., narrows the confidence interval, in the resulting GP model because the non-negativity information is incorporated.  While this paper covers only the non-negativity bound, other inequality constraints can be enforced in a similar manner. 

\section{Acknowledgement}
Xueyu Zhu's work was supported by Simons Foundation.
Xiu Yang's work was supported by the U.S. Department of Energy 
Office of Science, Office of Advanced Scientific Computing Research 
as part of Physics-Informed Learning Machines
for Multiscale and Multiphysics Problems (PhILMs).